\documentclass[eat,twocolumn]{jmlr} %

\usepackage{longtable}% for long tables

\usepackage{booktabs}
\usepackage[load-configurations=version-1]{siunitx} % newer version
\usepackage{amssymb}

\usepackage[font=small,skip=-2pt,belowskip=-10pt]{caption}
\theorembodyfont{\upshape}
\theoremheaderfont{\scshape}
\theorempostheader{:}
\theoremsep{\newline}

\jmlrvolume{}
\firstpageno{1}
\editors{}

\jmlryear{2022}
\jmlrworkshop{Machine Learning for Health (ML4H) 2022}

\title[Online MixEHR]{Phenotype Detection in Real World Data \titlebreak  via Online MixEHR Algorithm } 
 \author{\Name{Ying Xu} \Email{yxu@eqrx.com}\\
       \Name{Romane Gauriau} \Email{rgauriau@eqrx.com}\\
   \Name{Anna Decker} \Email{adecker@eqrx.com}\\
   \Name{Jacob Oppenheim} \Email{joppenheim@eqrx.com}\\
   \addr EQRx Inc. 50 Hampshire St, Cambridge, MA 02139}

\begin{document}

\maketitle

\begin{abstract}
Understanding patterns of diagnoses, medications, procedures, and laboratory tests from electronic health records (EHRs) and health insurer claims is important for understanding disease risk and for efficient clinical development, which often require rules-based curation in collaboration with clinicians. We extended an unsupervised phenotyping algorithm, mixEHR, to an online version allowing us to use it on order of magnitude larger datasets including a large, US-based claims dataset and a rich regional EHR dataset. In addition to recapitulating previously observed disease groups, we discovered clinically meaningful disease subtypes and comorbidities. This work scaled up an effective unsupervised learning method, reinforced existing clinical knowledge, and is a promising approach for efficient collaboration with clinicians. 

\end{abstract}

\section{Introduction}
\label{sec:intro}

Electronic health records (EHRs) and health insurer claims datasets are rich sources of patient information that provide opportunities to improve clinical decision-making, deliver healthcare more effectively, and improve efficiency of drug development. Modeling meaningful patterns in these types of data is a challenge. Latent Dirichlet Allocation (LDA) \citep{blei2003latent} has been widely used in natural language processing to discover common topics in collections of documents. \cite{nakamura2009grounding} extended LDA to multimodal categorization and words grounding for robots. \cite{li2020inferring} proposed an analogy between the automatic text categorization and uncovering underlying disease phenotypes from EHR data, considering each patient as a document, each disease meta-phenotype as a topic, and diagnosis, medications, and other codes as words. The proposed algorithm, mixEHR, demonstrated great success at revealing biologically meaningful topics from EHR data in an unsupervised setting. However, it did not scale very well to large state of the art datasets. Inspired by \cite{hoffman2010online}, we extended the mixEHR algorithm to an online version and implemented it under a popular Python topic model framework, Gensim \citep{rehurek_lrec}. We then applied it to two large datasets, including a claims dataset of over $30$ million patients and a regional EHR dataset of about 2 million patients, both from the United States. We found biologically meaningful disease groups and  sub-categories of diseases that reinforce the findings from \cite{li2020inferring} in a large and highly representative national dataset as well as a rich regional EHR dataset. 

%\textsf{epsfig}.\footnote{See
%\url{http://www.ctan.org/pkg/l2tabu}}

%\begin{note}

%\end{note}

\section{Online mixEHR algorithm}

Online LDA proposed by \cite{hoffman2010online} is based on online stochastic optimization and can handily analyze massive document collections, including those arriving in a stream in a more efficient way. Here we extended the algorithm to handle different data categories for the same document (see \figureref{fig:plate} in the Appendix for a graphical representation of the model) to obtain an online version of the mixEHR model. % \citep{li2020inferring}.

%The mixEHR model is a natural extension of the standard LDA model \citep{blei2003latent},  handling different data categories for the same document (see \figureref{fig:plate} in the Appendix for a graphical representation of the model).  Instead of a single $\beta$, we have a set of $\{\beta^1, \dots, \beta^T\}$ (where $T$ stands for the total number of data categories). Each data category has its own word distribution.

We used Variational Bayesian (VB) inference to approximate the true posterior by a simpler distribution $q(z^1,\dots, z^T, \theta, \beta^1,\dots, \beta^T)$. Using similar notations to \cite{hoffman2010online}, the Evidence Lower BOund (ELBO) is as follows:
\begin{small}
\begin{align*}
& \log p(w^1,\dots, w^T|\alpha, \eta^1, \dots, \eta^T) \\
 \geq & \mathcal{L}(w^1,\dots, w^T,\phi^1, \dots,\phi^T,\gamma, \lambda^1, \dots, \lambda^T) \\
  \overset{\Delta}{=} & \mathbf{E}_q[\log p(w^1,\dots, w^T, z^1,\dots, z^T, \theta, \\
  &\qquad \qquad \beta^1, \dots, \beta^T |\alpha, \eta^1, \dots, \eta^T) ] \\
 & \qquad - \mathbf{E}_q[\log q(z^1,\dots, z^t, \theta, \beta^1, \dots, \beta^T)].
\end{align*}
\end{small}

Similarly, we chose a fully factorized distribution $q$ of the form
\begin{small}
\begin{align*}
    q(z_{di}^t = k) &= \phi_{dw_{di}k}^t, \; t = 1, \dots, T; \\
    q(\theta_d) &= \text{Dirichlet}(\theta_d; \gamma_d); \\
    q(\beta_k^t) &= \text{Dirichlet}(\beta_k^t; \lambda_k^t), \; t = 1, \dots, T.
\end{align*}
\end{small}

Thanks to the factorization, the updates of $\mathbf{\phi}$ and $\mathbf{\lambda}$ remain the same. The update of $\gamma$ becomes:
\begin{small}
\begin{equation*}
    \gamma_{dk} = \alpha + \frac{1}{T} \sum_t \sum_w n_{dw}^t \phi_{dwk}^t,
\end{equation*}
\end{small}
which is the average updates among different data categories.

We then obtained an extension of the Algorithm 2 in \cite{hoffman2010online}.

\section{Results}
\label{sec:res}

We applied the online mixEHR algorithm to two datasets. The first was a medical and pharmacy claims dataset from Optum's de-identified \textsc{Clinformatics}$^{\circledR}$ Data Mart (CDM) Database. The CDM data were derived from administrative health claims for members of large commercial and Medicare Advantage health plans. The data contained information for approximately $30.5$ million patients in the United States. We jointly modeled three data categories: diagnosis and procedure codes, and medication orders/administrations. The other dataset was structured EHR data from Geisinger Health, a regional health care provider and payer covering approximately $2$ million patients. Although a regional dataset, Geisinger had high quality lab codes associated with patient encounters, allowing us to model laboratory data as well.
Results shown below are all from models using $110$ topics, which are chosen by comparing perplexity scores on validation datasets. 

\subsection{Clinically meaningful topics}

We examined all of the learned topics by qualitatively assessing the coherence of the top codes from each data category, and found that most of them were indeed clinically meaningful as shown in \figureref{fig:topics}. The words from different data categories reflected the same theme and strongly influenced the topic regardless of their dictionary sizes. Many topics focused on the same type of disease, such as the Breast cancer and Bipolar disorder topics. Others reflected more complex conditions, e.g. \figureref{fig:topics}, `CKD with T2D'. This topic was a subset of chronic kidney disease (CKD) concurrent with Type 2 diabetes (T2D). The top medication, losartan, is used to treat hypertension and to help protect the kidneys from damage due to diabetes.  These disease categories arose naturally from the probabilistic generating process of mixEHR and were not reflected in any data modality individually.

\begin{figure}[htbp]
\floatconts
  {fig:topics}
  {\caption{ Example topics derived from the Optum claim dataset. The x-axis represents the word probability of belonging to the topic and the y-axis represents the descriptions of the top codes used. Finally, the color denotes the data type, and the words are ranked within each data type in the figure.}} %\protect\footnotemark. }}
  {   \includegraphics[width=\linewidth]{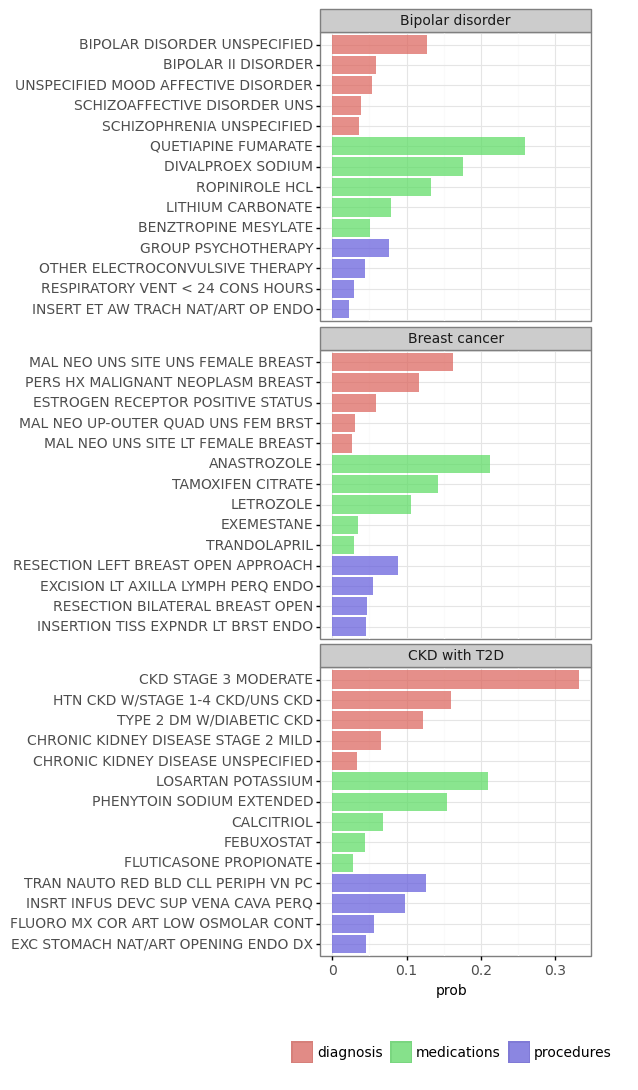}}
\end{figure}

%\footnotetext{The y-axis shows the descriptions of top codes within each topic and the x-axis represents the inferred probabilities of these codes belonging to each topic.}

\subsection{Sub-types of disease}

For common diseases with large populations, we obtained multiple related topics. Upon careful examination, we found that they are not simply repeating themselves, but reflected different sub-types of the same disease. For example, the two topics shown in \figureref{fig:T2D} were both topics of Type 2 Diabetes. The key differences were found in medication usage, where the topic in `T2D-a' used metformin, and the topic in `T2D-b' mainly used insulin, likely reflecting disease stage and severity.

\begin{figure}[htbp]
\floatconts
  {fig:T2D}
  {\caption{Sub-types of Type 2 Diabetes derived from the Optum claims data.}}
  {%
   % \subfigure[T2D-a]{\label{fig:T2D-a}%
      \includegraphics[width=\linewidth]{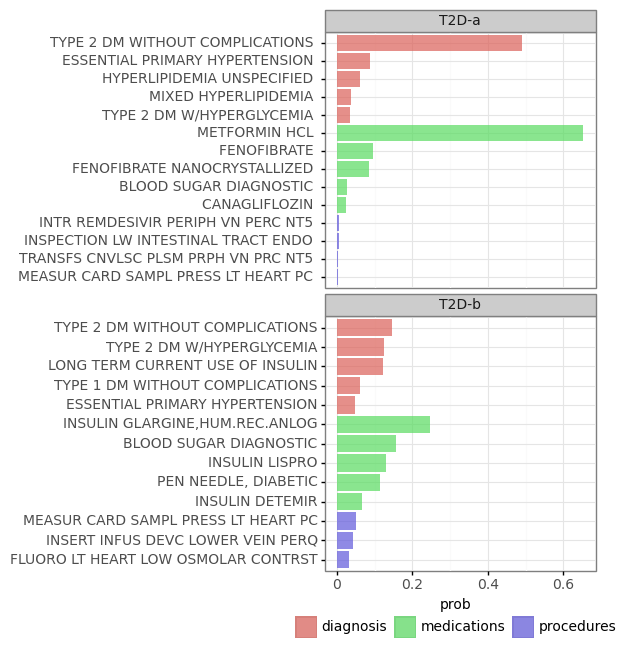}}%
  
\end{figure}

\subsection{Disease comorbidity}

In addition to discovering clinically meaningful topics, we also discovered some interesting connections between different diseases. For example, in our Multiple myeloma (MM) topic (\figureref{fig:mm}), one of the top medications, aciclovir, is used as prophylaxis for MM patients receiving Velcade (bortezomib) for herpesvirus infections. This co-occurrence of MM and herpes was confirmed both in the literature \citep{rettig1997kaposi, kastrinakis2000molecular, zheng2022efficacy} and clinicians  who we consulted. However, the study sample sizes in the literature were limited, where there were only $10$ non-diseased individuals and $16$ in \cite{rettig1997kaposi} and 719 MM patients in \cite{zheng2022efficacy}. Our findings from an unsupervised approach based on EHR data without highly specific curation have enriched the evidence of this association in a general, representative population. 

\begin{figure*}[h]
\floatconts
  {fig:mm}
  {\caption{Multiple myeloma and herpes derived from the Geisinger EHR data.}}
  {    \includegraphics[width=0.85\linewidth]{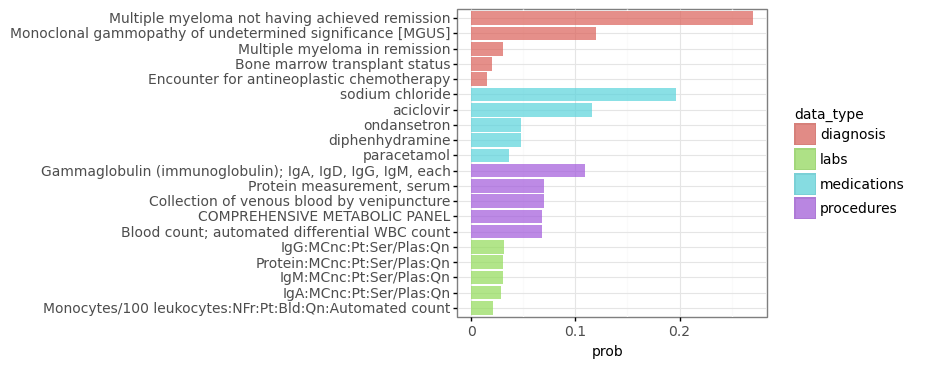}}%
\end{figure*}

\subsection{Cohort separation}

Identifying precise cohorts of patients with specific medical conditions has many use cases, such as clinical trial recruitment, outcome prediction, and as a starting point for retrospective studies  \citep{shivade2014review}. We defined several specific disease cohorts in collaboration with clinicians in the EHR data based on combinations of repeated diagnosis codes, medications, confirmatory procedures and lab results. Then, we examined whether the mixEHR embedding could distinguish different cohorts.

\figureref{fig:cht} shows the topic loading distribution within each cohort. In general, we observed relatively clear separation of cohorts by individual topics. The Ulcerative colitis (UC) cohort did not have much overlap with others and concentrated in topic $39$ (gastrointestinal disorders). The Atopic dermatitis (AD) cohort was mainly in topic $78$ (dermatitis). Rheumatoid arthritis (RA) patients had strong loading in topic $44$. Psoriatic arthritis (PsA) and Axial spondyloarthritis (AxSpA) both had concentrations in topic $44$, which is not surprising as they are all types of autoimmune arthropathies. Differences in other topic signatures could still differentiate them. We observed our PsA cohort split between several topics: one group ($63$,$88$) representing Psoriasis, the other ($44$) severe PsA. We confirmed with clinicians that this is reflective of etiology of PsA, which develops from Psoriasis. Topic $92$ was focused on Breast cancer (BC) and both BC and metastatic Breast cancer cohorts had high loading in it. Late stage metastatic BC was identified with topic $45$, focused on cancers in other locations, which is consistent with the definition of metastasis.

\begin{figure*}[h]
\floatconts
  {fig:cht}
  {\caption{Topic loading distributions for different cohorts using the Geisinger EHR data. Using the mixEHR model we predicted the topic loading for each patient within cohorts of interest. Then, we plotted boxplots of the topic loading distributions for each cohort. Here we showed only the relevant topics  on the x-axis for each cohort to preserve space.}}
  {    \includegraphics[width=0.85\linewidth]{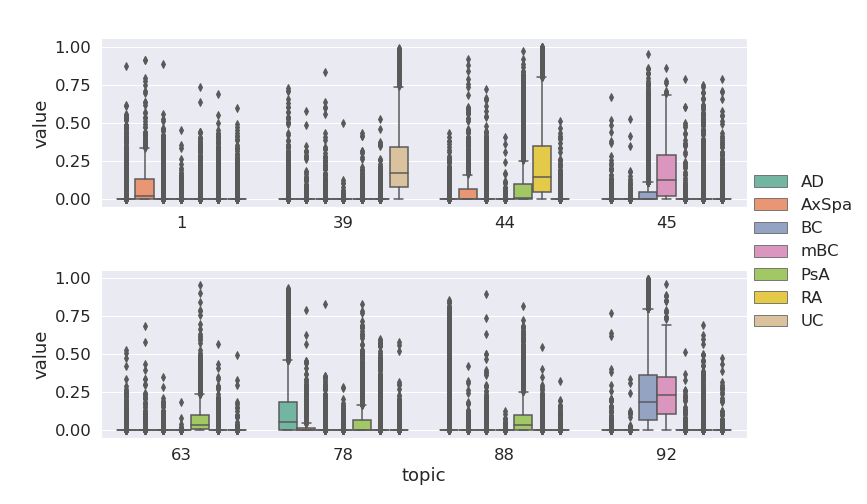}}%
\end{figure*}
\vskip20mm

\subsection{Identifying synonymous codes}

The  Geisinger EHR dataset uses LOINC (Logical Observation Identifiers Names and Codes) codes to classify labs, which contain very detailed information such as compnent, time, scale, specimen, and method about each lab taken. Thus, multiple codes are often generated for one test. For example, for a standard blood panel, multiple components are measured, such as red blood cells, white blood cells and hemoglobin, the data contain one code for each component. The mixEHR algorithm provides a natural way to obtain these groupings automatically.

For each lab code $i$, denote $\hat{p_i}$ as  the topic probability distribution learned from the mixEHR model. Thus, each lab code is represented by a $k$ dimensional vector, where $k$ is the number of the topics. Then using the cosine similarity we can define the similarity between the code $i$ and code $j$ as follows:
\begin{equation*}
 \frac{\hat{p_i}\cdot\hat{p_j}}{ \|\hat{p_i}\| \|\hat{p_j}\|}.    
\end{equation*}

Once we obtained the similarity matrix, we can then find highly related code groups. \figureref{fig:similarity} illustrates the cosine similarity matrix of the lab codes after re-ordering by groups, which shows a clear block diagonal pattern.    
\begin{figure}[h]
\floatconts
  {fig:similarity}
  {\caption{The cosine similarity matrix of the lab codes ordered by groups. The brighter the color the higher similarity is.}}
  {\includegraphics[width=\linewidth]{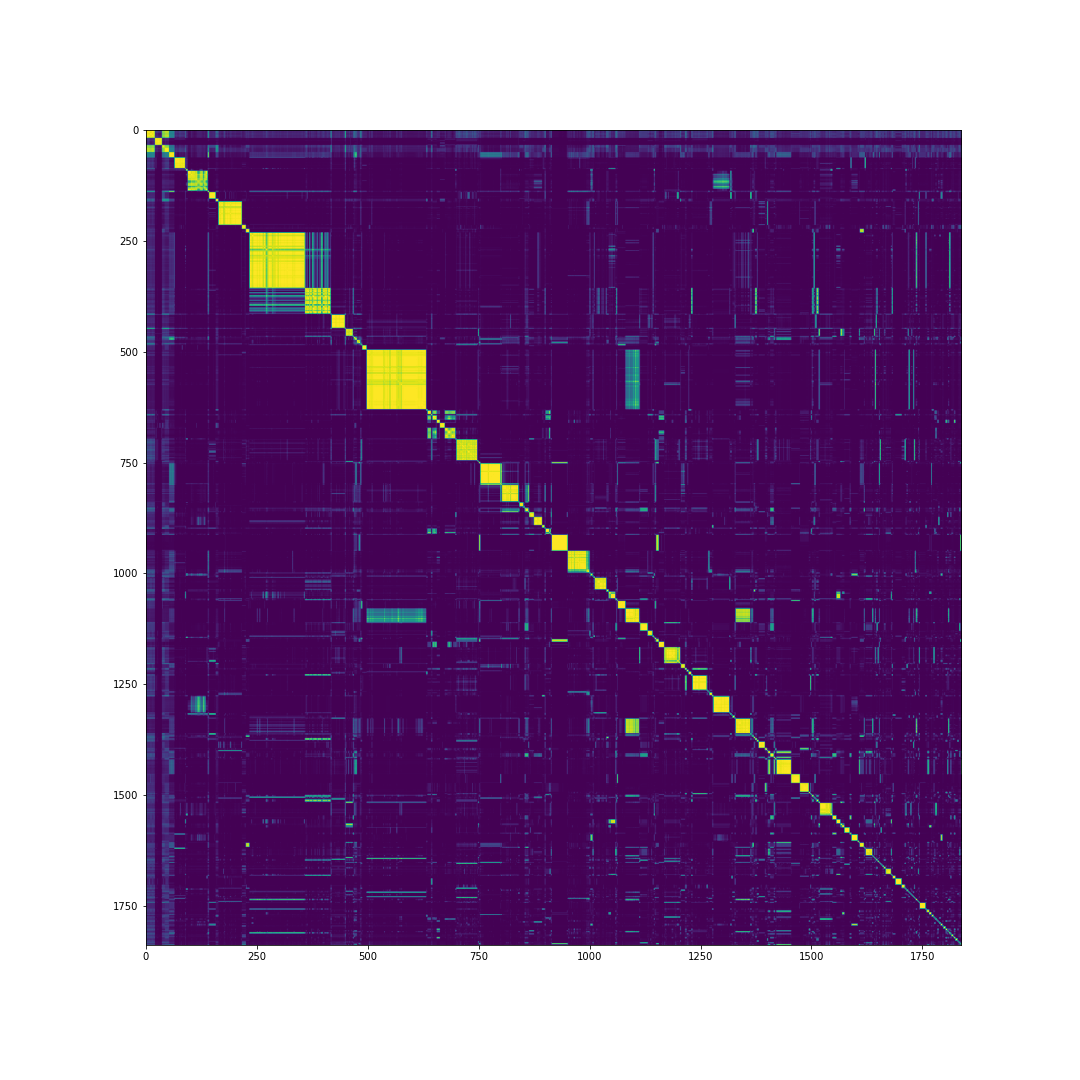}}
\end{figure}

Upon examination some blocks are effectively the  same test with different measurement units, others correspond to different components from the same test, and some might be different  tests related to the same topic, which require further validation by domain experts. Two example groups are illustrated in Table \ref{tab:bld} and Table \ref{tab:allergy}. Codes in Table \ref{tab:bld} all correspond to the standard blood test while in Table \ref{tab:allergy} are all different allergy tests.

\vskip5mm
\begin{table}[h] \centering 
  \caption{Learned groupings of blood tests based on cosine similarity} 
  \label{tab:bld} 
  \tiny
\begin{tabular}{@{\extracolsep{5pt}} ll} 
\\[-1.8ex]\hline 
\hline \\[-1.8ex] 
Code & Description \\ 
\hline \\[-1.8ex] 
C0362892 & Basophils:NCnc:Pt:Bld:Qn:Automated count \\ 
C0362894 & Basophils/100 leukocytes:NFr:Pt:Bld:Qn:... \\ 
C0362900 & Eosinophils:NCnc:Pt:Bld:Qn:Automated count \\ 
C0362902 & Eosinophils/100 leukocytes:NFr:Pt:Bld:Qn:... \\ 
$\cdots$ & $\cdots$ \\
 C0366777 & Hematocrit:VFr:Pt:Bld:Qn:Automated count \\ 
C0362923 & Hemoglobin:MCnc:Pt:Bld:Qn \\ 
C0484430 & Leukocytes:NCnc:Pt:Bld:Qn:Automated count \\ 
C0362952 & Lymphocytes/100 leukocytes:NFr:Pt:Bld:Qn:... \\ 
C0362960 & Monocytes/100 leukocytes:NFr:Pt:Bld:Qn:... \\ 
 C0362987 & Neutrophils/100 leukocytes:NFr:Pt:Bld:Qn:... \\ 
 C0362994 & Platelets:NCnc:Pt:Bld:Qn:Automated count \\ 
\hline \\[-1.8ex] 
\end{tabular} 
\end{table} 

\vskip5mm
\begin{table}[h] \centering 
  \caption{Learned groupings of allergy tests based on cosine similarity} 
  \label{tab:allergy} 
  \tiny
\begin{tabular}{@{\extracolsep{5pt}} ll} 
\\[-1.8ex]\hline 
\hline \\[-1.8ex] 
Code & Description \\ 
\hline \\[-1.8ex] 
C0484008 & Honey Ab.IgE:ACnc:Pt:Ser:Qn \\ 
C0483797 & Prunus avium Ab.IgE:ACnc:Pt:Ser:Qn \\ 
C0365326 & Cardiolipin Ab.IgM:ACnc:Pt:Ser:Qn:IA \\ 
C0483951 & Agrostis stolonifera Ab.IgE:ACnc:Pt:Ser:Qn \\ 
C0362651 & Piper nigrum Ab.IgE:ACnc:Pt:Ser:Qn \\ 
C0483715 & Apis mellifera Ab.IgE:ACnc:Pt:Ser:Qn \\ 
C1976891 & Wine vinegar Ab.IgE:ACnc:Pt:Ser:Qn \\ 
$\cdots$ & $\cdots$ \\
C0362644 & Fagus grandifolia Ab.IgE:ACnc:Pt:Ser:Qn \\ 
C0362889 & Dolichovespula arenaria Ab.IgE:ACnc:Pt:Ser:Qn \\ 
C0484283 & Zizania spp Ab.IgE:ACnc:Pt:Ser:Qn \\ 
\hline \\[-1.8ex] 
\end{tabular} 
\end{table} 

Another use case is to build bridges between different medical coding systems. In the EHR or health insurer claims system, it is common to see multiple coding systems been used. For example, for medical procedures, ICD-10-PCS \footnote{The International Classification of Diseases, Tenth Revision, Procedure Coding System}, CPT \footnote{Current Procedural Terminology} and HCPCS \footnote{Healthcare Common Procedure Coding System} are commonly used, but there is no mapping between them. This approach could provide a starting point for such a map.

\section{Discussion}

In this work, we were able to scale up the original mixEHR algorithm,  enabling adaptation to new data without refitting to the entire dataset. Similar to previous work, we were able to learn relevant disease topics directly from data without supervised training, enabling fruitful collaboration with clinicians. Additionally, our findings in large claims and EHR datasets confirmed at a large scale findings from previous case reports, such as the co-occurrence between multiple myeloma and acyclovir for herpesvirus, and uncovered potential indicators of disease progression and disease sub-types, such as different T2D treatment profiles. Interestingly, the results in claims data were clinically meaningful even though we did not include lab data. 

One key aspect of topic modeling is choosing the number of topics. Here, we presented results for $110$ topics for both claims and EHR. We experimented with different numbers of topics and found that within a range, the perplexity scores were not significantly different. As the number of topics increased, single-disease focused topics tended to appear. However, this was not a hard and fast rule. We saw common disorders such as T2D span multiple topics reflecting underlying severity. We also identified stages of PsA progression from Psoriasis to full-blown PsA, reflecting well-known clinical trajectories. In contrast, RA remained one topic no matter the amount of data despite relatively high prevalence, breadth of severity, and types of treatment. Thus, the relationship between data volume, noise, and number of topics remains an open question. 

These results demonstrated the power of unsupervised methods training across data modalities to infer clinically relevant disease topics. Working directly with clinicians and identifying the optimal scale for topic definitions and best use cases are next steps. Accommodating patient level information is also of interest, but increasing model complexity versus increasing data size remains to be investigated.

%\acks{Acknowledgements go here.}

%\newpage
\bibliography{pmlr-sample}

%\appendix

\section*{Appendix}\label{apd:first}
MixEHR can be represented as a probabilistic graphical model as in \figureref{fig:plate}. 
\begin{figure}[htbp]
\floatconts
  {fig:plate}
  {\caption{Graphical model representation of MixEHR. The boxes are ``plates'' representing replicates. The outer plate represents documents and the inner plate represents the repeated choice of topics and words within a document. }}
  {\includegraphics[width=0.9\linewidth]{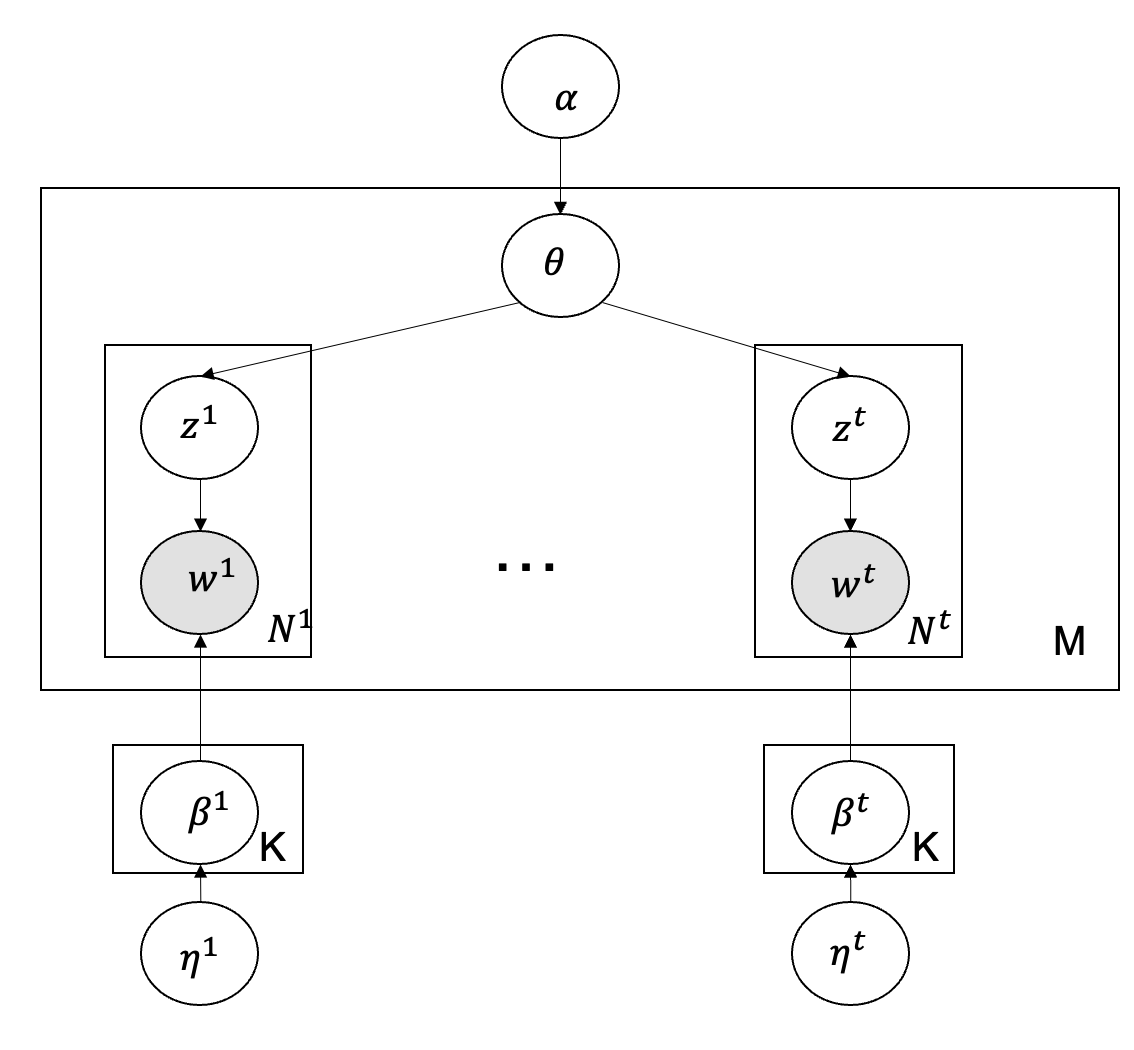}}
\end{figure}

It assumes a collection of $K$ ``topics''. Each topic defines a multinomial distribution over each data category's  vocabulary and is assumed to have been draw from a Dirichlet, $\beta^{t} \sim \text{Dirichlet}(\eta^{t})$. Given the topics, mixEHR assumes the following generative process for each document:
\begin{enumerate}
    \item Draw a distribution over topics $\theta \sim \text{Dirichlet}(\alpha)$
    \item Within each data category $t$, for each word $w^{t}$, draw a topic index $z^{t} \sim \text{Multinomial}(\theta)$
    \item Then given $z^t_n, \beta^t$ draw $w_n^t$ from a multinomial distribution conditioned on the topic $z^t_n$
\end{enumerate}

\vspace{3mm}
The size of each data modality/vocabulary is shown in Table \ref{tab:dim}. There is substantial variability across modalities. 

\vspace{3mm}
\begin{table}[hbtp]
\small
\floatconts
  {tab:dim}
  {\caption{The vocabulary sizes of datasets}}
  {\begin{tabular}{lrr}
  \toprule
  \bfseries Data category & \bfseries Optum & \bfseries Geisinger\\
  \midrule
  Diagnosis codes & 66134 & 44174 \\
  Medication codes & 6062 & 1948\\
  Procedure codes & 27227 & 25758\\
  Lab codes & NA & 3750 \\
  \bottomrule
  \end{tabular}}
\end{table}

\end{document}